\documentclass[times, 10pt,twocolumn]{article}
\usepackage{spconf}
\usepackage{epsfig}
\usepackage{graphicx}
\usepackage{graphics}
\begin{document}
\title{Learning Isometric Separation Maps}

\name{Nikolaos Vasiloglou, Alexander G. Gray, David V. Anderson \thanks{This work was sponsored by Google grants}}
\address{Georgia Institute of Technology\\
         Atlanta GA 30332\\
        nvasil@ieee.org, agray@cc.gatech.edu, dva@ece.gatech.edu\\
}

\maketitle
\thispagestyle{empty}
\begin{abstract}
Maximum Variance Unfolding (MVU) and its variants have been very
successful in embedding data-manifolds in lower dimensional
spaces, often revealing the true intrinsic dimension. In this paper
we show how to also incorporate supervised class information into an
MVU-like method without breaking its convexity. We call this method
the Isometric Separation Map and we show that the resulting kernel
matrix can be used as a binary/multiclass Support Vector Machine-like method in
a semi-supervised (transductive) framework. We also show that the
method always finds a kernel matrix that linearly separates the
training data exactly without projecting them in infinite
dimensional spaces. In traditional SVMs  we choose a kernel and hope that the data become linearly separable in the kernel space. In this paper we show how the hyperplane can be chosen ad-hoc and the kernel is trained so that data are always linearly separable. Comparisons with Large Margin SVMs show comparable performance.
\end{abstract}

\section{Introduction}
\label{sec_introduction} Support Vector Machines have been quite
successful in separating classes of data that are not linearly
separable. The kernel trick lifts the data in a high dimensional
Hilbert space usually of infinite dimension
\cite{shawetaylor2004kmp}. Embedding datasets in infinite
dimensional spaces gives the advantage  of separating data with linear
hyperplanes in the lifted space, that otherwise were not separable
in the original space. So far it is not clear how the dimensionality
of the kernel affects the performance of SVMs. It is not known yet
how many dimensions are sufficient for separating the classes. It is
very likely that the minimum dimension required for linear
separability is much smaller than the original dimension of the
data. This is because the data might already be embedded in a
manifold with redundant dimensions.

Maximum Variance Unfolding (MVU) \cite{weinberger2004lkm} along with
other manifold learning methods has addressed the problem of
reducing the dimensionality of the data by preserving local
distances. Most of the time the data end up living in a lower
dimensional space. MVU explicitly finds the optimal kernel matrix
for the data, by solving a semidefinite program. As a remark MVU
usually gives the most compact spectrum \cite{weinberger2004lkm}, revealing the true intrinsic dimensionality of the dataset very
well. The authors of the MVU point out though that it has very poor
performance when it comes to using the kernel matrix for SVM
classification \cite{weinberger2004lkm}  as it does not include any
information about the linear separability of the classes. For
example in figure \ref{fig_two_class_swiss_roll}b we show two
classes on a Swiss roll manifold. After unfolding with MVU
\ref{fig_two_class_swiss_roll}d, the classes remain non-linearly
separable.

In this paper we introduce a variation of MVU that takes into
consideration the linear separability of the classes. The result is
a new algorithm, the Isometric Separation Mapping (ISM), that gives an
unfolding that preserves the class structure of the manifold. The
algorithm can be seen as a transductive (semi-supervised SVM), since
it requires the test data during training. Previous work on
transductive SVMs has also been studied by several researchers. When
the choice of the kernel is ad-hoc, the problem becomes very
difficult as it boils down to mixed integer programming
\cite{bennett1999sss}. In \cite{lanckriet2004lkm} and
\cite{kim2008lkv} the authors train the kernel matrix over a set of
predefined kernels. Although this gives higher flexibility in
forming the kernel, it might still require a large number of
predefined kernels. For example if one of the choices was the
Gaussian, it would be necessary to keep a large number of them with
different sigmas (bandwidths). It is widely known that if the
bandwidth of the Gaussian is too wide or too narrow kernel methods
perform poorly. This technique usually leads to full rank Semidefinite programs that are computationally hard. Finally, in \cite{belkin2005mr} the Laplacian
Eigenmap framework is used for training SVMs. Laplacian Eigenmaps are another dimensionality reduction method based on the Gaussian kernel. It also tries to capture the local geometry and take advantage of it in SVM training. Our technique does not
make any assumption on the kernel function. The only requirement is
to preserve isometry on the data.

The paper is organized in the following way.
In section \ref{sec_MVU} we give an overview of MVU along with its
variants that make it scalable. In section \ref{sec_ISM} we
present the ISM algorithm. Some examples on  embedding manifolds with ISM are presented in \ref{sec_Dimensionality_Adjustment}. In section \ref{sec_TSVM} we
present a transductive SVM based on the ISM.

\section{Maximum Variance Unfolding, Maximum Furthest Neighbor Unfolding.}
\label{sec_MVU} Weinberger formulated the problem of isometric
unfolding as a Semidefinite Programming algorithm
\cite{weinberger2004lkm}. 

Given a set of data $X \in \Re^{N\times d}$, where $N$ is the number
of points and $d$ is the dimensionality, the dot product or Gram
matrix is defined as $G=X X^T $. The goal is to find a new Gram
matrix $K$ such that $rank(K)<rank(G)$ in other words $K=\hat{X}
\hat{X}^T$ where $\hat{X} \in \Re^{N\times d'}$ and $d'<d$.  Now the
dataset is represented by $\hat{X}$ which has fewer dimensions than
$X$. The requirement of isometric unfolding is that the euclidian
distances in the $\Re^{d'}$ for a given neighborhood around every
point have to be the same as in the $\Re^d$. This is expressed in:
\[
 K_{ii}+K_{jj}-K_{ij}-K_{ji}= G_{ii}+G_{jj}-G_{ij}-G_{ji}, \forall i, j \in I_i
\label{isometry}
\]
where $I_i$ is the  set of the indices of  the neighbors of the
$ith$ point. From all the $K$ matrices MFNU chooses the one that
maximizes the distances between furthest neighbor pairs and MVU the
one that maximizes the variance of the set (equivalently the
distances of the points from the origin). So the algorithm is
presented as an SDP:
\begin{eqnarray}
\label{MFNU}
   \max_{K} \;\sum_{i=1}^{N}B_{i} \bullet K &  &  \\
\nonumber   \mbox{subject to}  & & \\
\nonumber   A_{ij} \bullet K &=& d_{ij} \quad \forall j \in I_i  \\
\nonumber    K &\succeq& 0
\end{eqnarray}
where the $A \bullet X=\mbox{Trace}(AX^T)$ is the dot product
between matrices.
$A_{ij}$ has the following form:
\begin{equation}
\label{A_structure} A_{ij} = \left[
\begin{array}{cccccc}
  1 & 0 & \dots & -1 & \dots & 0 \\
  0 & \ddots & 0 & \dots & 0 & 0 \\
  \vdots & 0 & \ddots &  0 & \dots & 0 \\
  -1  & \dots & 0 & 1 & \dots & 0 \\
  \vdots & 0 & \dots & 0 & \dots & 0 \\
  0 & \dots & \dots & 0 &\dots & 0
\end{array}
\right]
\end{equation}
and
\begin{equation}
d_{ij}=G_{ii}+G_{jj}-G_{ij}-G_{ji}
\end{equation}
$B_i$ has the same structure of $A_{ij}$ and computes the distance $d_{ij}$
of the $i_{th}$ point with its furthest neighbor for MFNU, while for
MVU it is just the unit matrix (computes the distance of the points
from the origin).
 The last condition is just a centering constraint for the
covariance matrix. The new lower dimensional representation of data  $\hat{X}$ is found in the eigenvectors of $K$. In general MVU/MFNU gives Gram matrices that have compact
spectrum, at least more compact than traditional linear Principal
Component Analysis (PCA). The method behaves equally well with MVU.
Unfortunately this method can handle datasets of no more than
hundreds of points because of its complexity.

\subsection{The Non Convex MVU/MFNU.}
\label{NCMVU}
Kulis and Vasiloglou showed how the algorithm can be more scalable  \cite{kulis2007flr, vasiloglou2008ssm} by replacing the constraint $K\succeq 0$  \cite{burer2003npa} with
an explicit rank constraint $K=RR^T$. The problem becomes non-convex
and it is reformulated to:
\begin{eqnarray}
\label{MFNU_NC}
\max_{R} & &  \sum_{i=1}^{N} B_{i} \bullet RR^T \\
\nonumber& & \mbox{subject to:} \\
\nonumber & & A_{ij} \bullet RR^T=d_{ij}
\end{eqnarray}
In \cite{burer2003npa}, Burer proved that the above formulation has
the same global minimum with the convex one. In this form the
algorithm scales better. 

The above problem can be solved with the augmented Lagrangian method \cite{burer2003npa}.
\begin{eqnarray*}
\mathcal{L}= -B_{i} \bullet RR^T -\sum_{i=1}^{N}\sum_{\forall j \in I_i}
\lambda_{ij}(A_{ij}\bullet RR^T-d_{ij})+ \\
\frac{\sigma}{2}\sum_{i=1}^{N}\sum_{\forall j \in I_i}
\lambda_{ij}(A_{ij}\bullet RR^T-d_{ij})^2
\end{eqnarray*}
Our goal is to minimize the Lagrangian; that's why the objective
function is $-B_{i} \bullet RR^T$ and not $B_{i} \bullet RR^T$
The solution is typically found with the LBFGS method \cite{burer2003npa}.

\section{Isometric Separation Maps (ISM)}
\label{sec_ISM}

Although MVU and its variant MFNU give low rank kernel matrices,
experiments \cite{weinberger2004lkm} show
that they are performing  poorly when it comes to SVM
classification. In this section we will show that MVU/MFNU can be
modified so that the kernel matrix can be used for classification
too.

In traditional SVMs the kernel is chosen ad-hoc and the goal is to
find a hyperplane that can linearly separate the classes. The kernel
is chosen in such a way that it lifts the data in a high dimensional
space hoping that data would be linearly separable. In our approach
we have the hyperplane given and we are trying to find the kernel
matrix that separates the data along the hyperplane. Finding a
kernel matrix to satisfy that condition is trivial as it suffices to add one extra
dimension on the data that will be either -1 or 1. What is sort of
interesting though is to find a mapping  to a (higher or lower)
dimensional space that keeps data points linearly separable and
preserves the local isometry. As we will see later, depending on the
structure of the classes it is likely  to end up in a higher
dimensional space. We are interested in the minimum dimension of
that space.

The solution of the problem is the following. We pick one of the
data points $x_A$ to be normal to the separating hyperplane. The choice of the point
does not matter since it will just change the orientation of the
points in space. The manifold consists of two classes $C_1$ and $C_2$. Let $x_i \in C_1$ be the points that belong in the same class with $x_A$, then $k(x_A, x_i) \geq 0$, where $k(x_A, x_i)$ is the generalized dot product between $x_A$ and $x_i$. For points
that belong to the opposite class $x_i \in C_2$, $k(x_A, x_i) \leq
0$. Now the problem of MVU/MFNU  with linear separability
constraints can be cast as the following Semidefinite Program:
\begin{eqnarray}
\label{LSMFNU}
   \max_{K} \;\sum_{i=1}^{N}B_{i} \bullet K &  &  \\
\nonumber   \mbox{subject to}  & & \\
\nonumber   A_{ij} \bullet K &=& d_{ij} \quad \forall j \in I_i  \\
\nonumber   K_{A,i} &\geq& 0, \quad  \forall i \in C_1\\
\nonumber   K_{A,i} &\leq& 0, \quad  \forall i \in C_2\\
\nonumber    K &\succeq& 0
\end{eqnarray}

Using the same formulation as in \cite{vasiloglou2008ssm} we can
solve the above problem in a non-convex framework that scales
better. Extending the problem for more classes is pretty straight
forward. The only modification is to use more anchor points that
will serve as normal vectors to the separating hyperplanes.  The problem is always feasible provided that $k\ll N$. \footnote{As long as the k neighbors belong to the tangent space and the manifold is smooth, a folding (locally isometric transform) of the
manifold along a hyperplane always exists \cite{lee2003ism}}. If all
pair distances are given then the Gram matrix  is uniquely defined
and the problem might be infeasible. In the trivial case where $k=1$ meaning that each point has exactly one neighbor then the problem is always feasible. In general there is a maximum k where the problem might become infeasible. That means there is always a $k$ where the training error is zero. That means we can always find a dimensional space where the Manifold can be embedded isometrically.

If some of the data points are labeled (training data) and some are not (test data), then the above method can be used as an SVM-like classifier that always achieves zero training error in contrast to other algorithms proposed for learning the kernel in SVM (mentioned in
section \ref{sec_introduction} , where the kernel is learnt as a
convex combination of preselected kernels).
This might sound as over-fitting on the training data. In reality though this is not true since the test data participate during training glued on the training data with the distance constraints. Another remark on the
ISM is that it is not a max margin classifier because it does not
regularize the norm of the normal vector. It is not possible to do
it since we need to also preserve the local distances. 


\section{Dimensionality Minimization with ISM}
\label{sec_Dimensionality_Adjustment} 
\begin{figure}[!h]
\centerline{a\includegraphics[height=3.0cm]{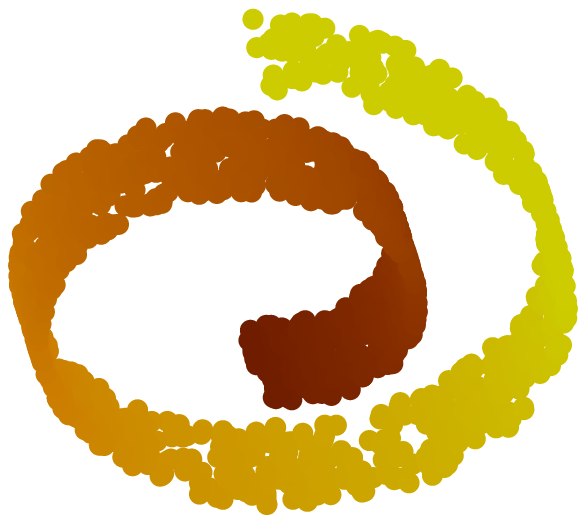}
b\includegraphics[height=2.0cm]{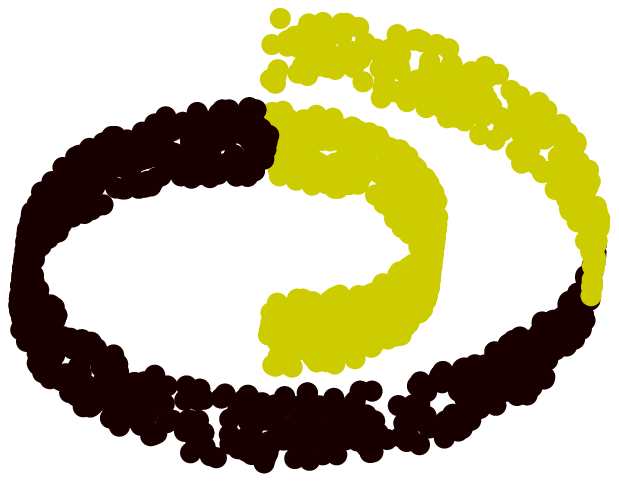}}
\centerline{c\includegraphics[height=3.0cm]{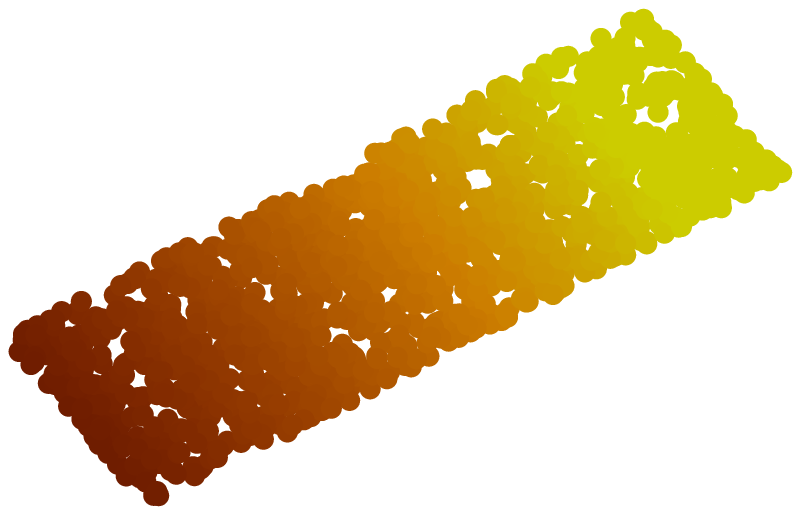}
d\includegraphics[width=2.0cm]{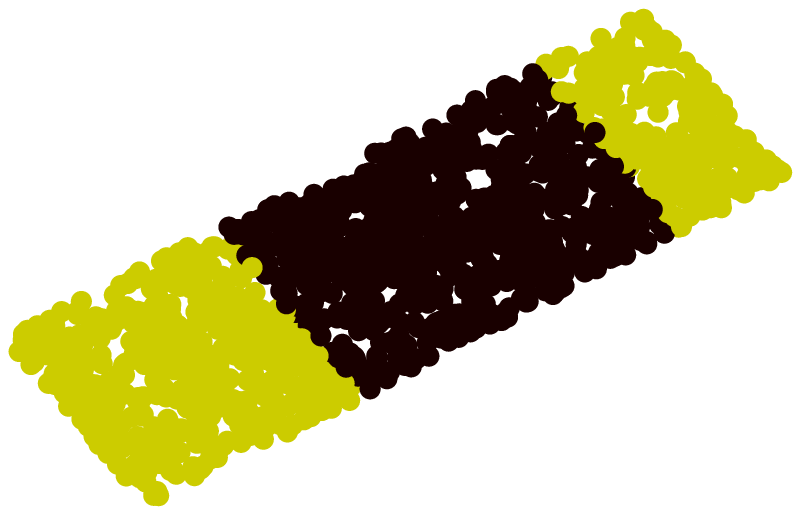}}
\centerline{e\includegraphics[height=4.0cm]{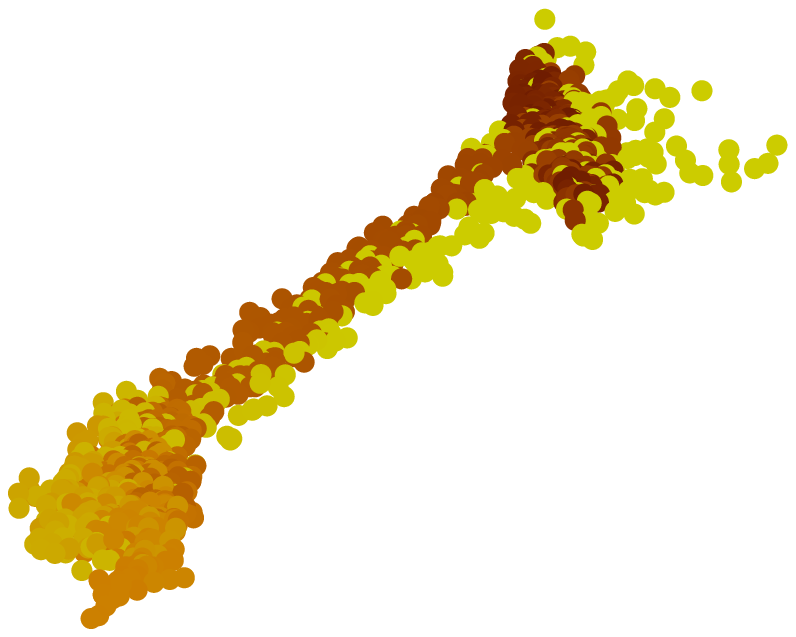}
f\includegraphics[height=2cm]{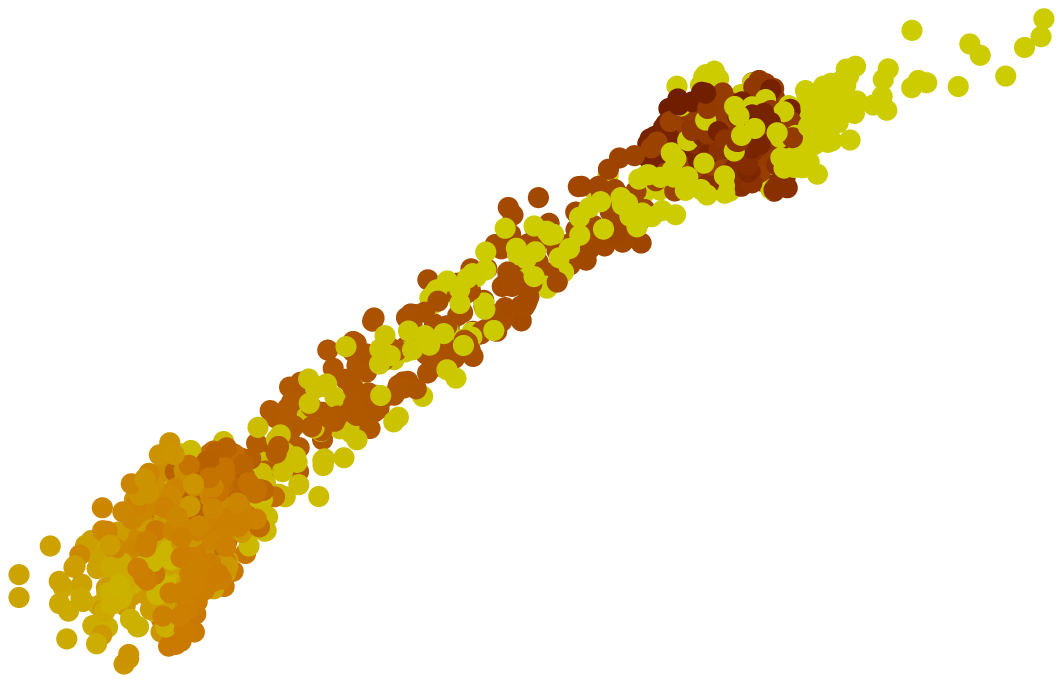}}
\centerline{g\includegraphics[height=4cm]{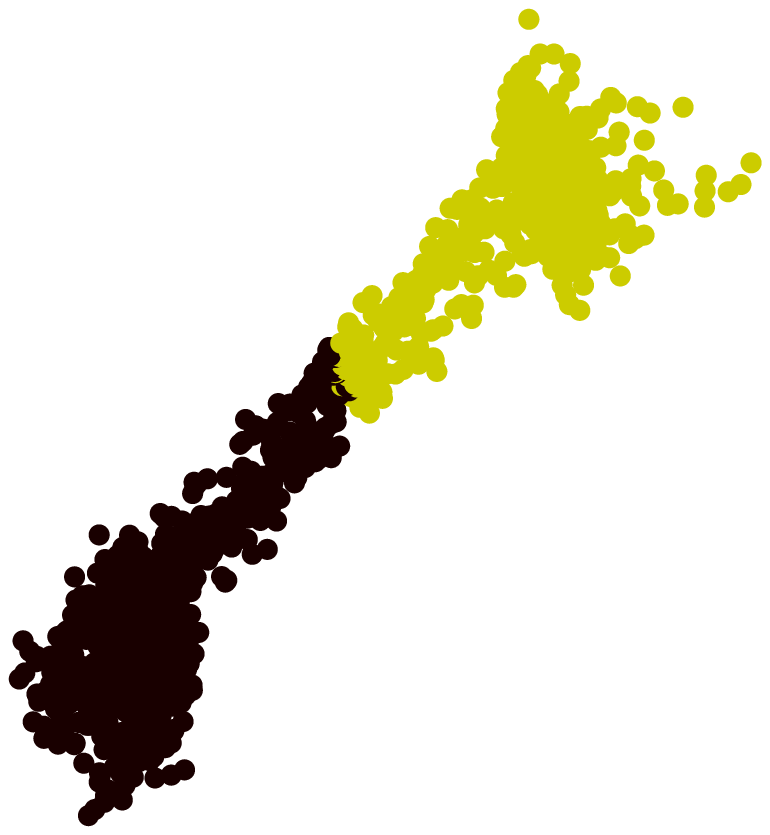}
h\includegraphics[height=2cm]{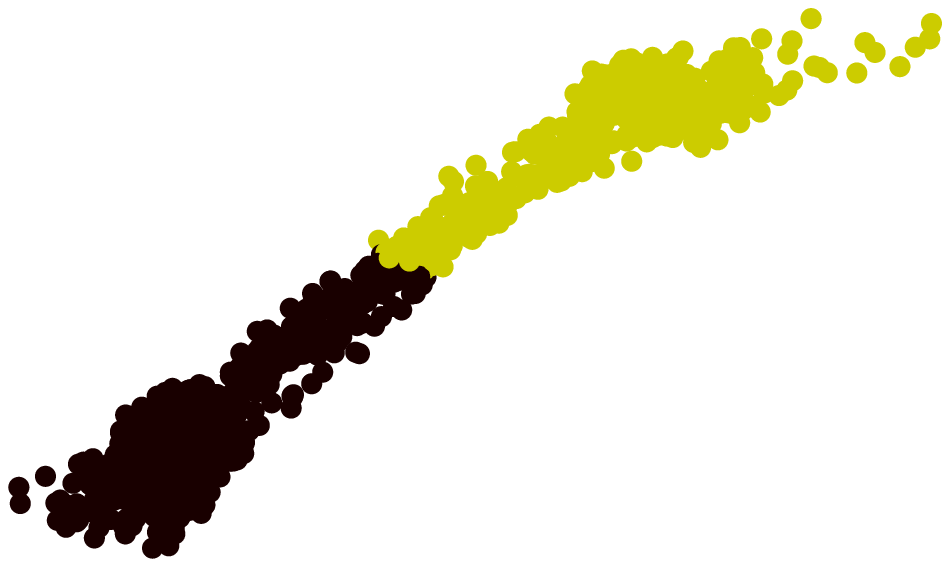}}
\caption{a)A three dimensional swiss roll painted with color
gradient. b)The same swiss roll with two classes on it, black and
green c)Unfolded swiss roll (a) with MVU/MFNU (no class
information). The color gradient shows that local distances has been
preserved. d) Unfolded swiss roll (b) with MVU/MFNU. The two classes
are not linearly separable. e,f) Views of the swiss roll (a) with
ISM. The class structure was taken from (b). The intension of this
figure is to show how the points are mapped so that the local
neighborhoods are preserved. g,h)Views of the (b) manifold after
ISM. Now points are painted with the class colors to show that they
are linearly separable}
\label{fig_two_class_swiss_roll}
\end{figure}

\begin{figure}[!h]
\centerline{\includegraphics[height=2.0cm]{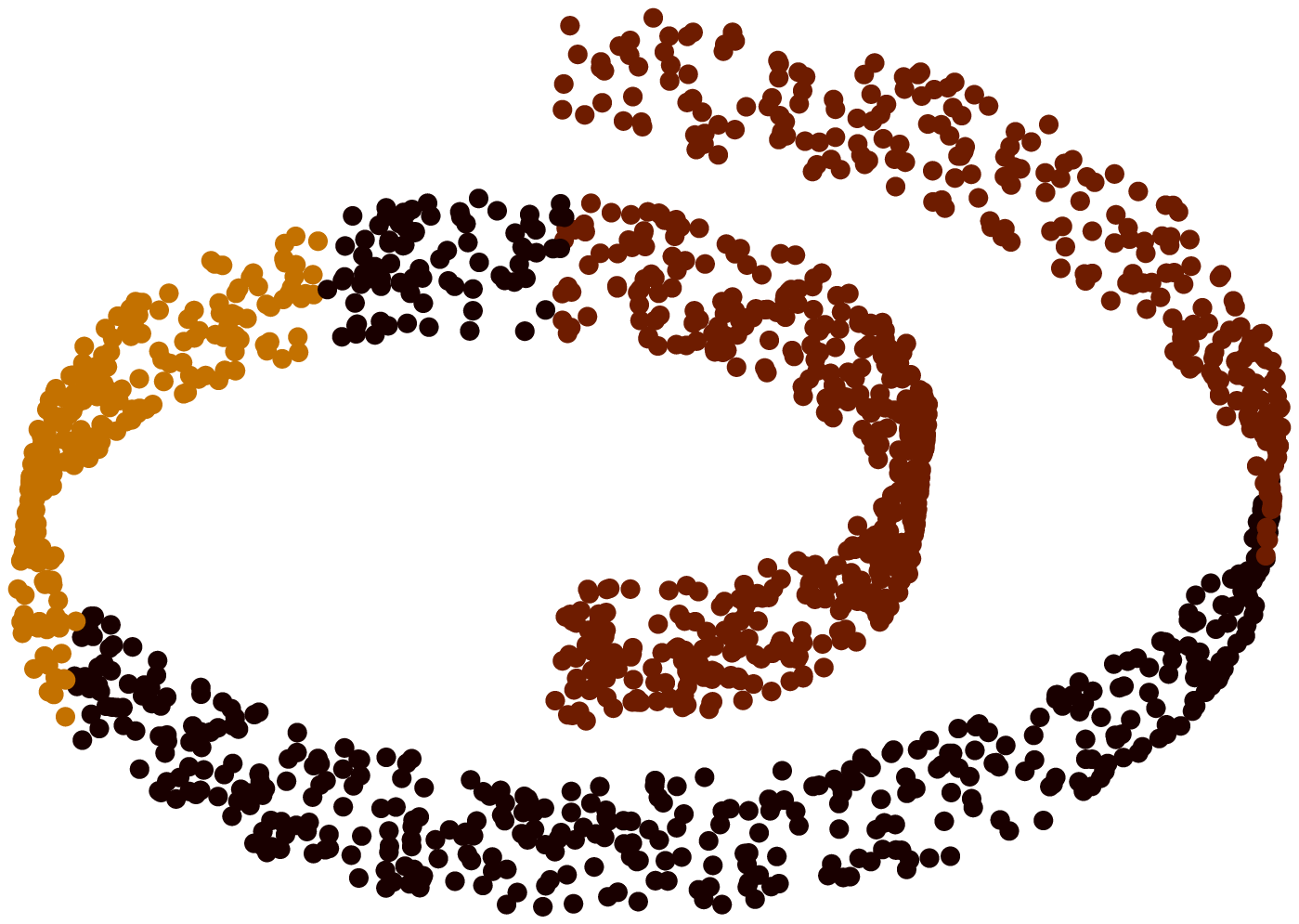}
            \includegraphics[height=2.0cm]{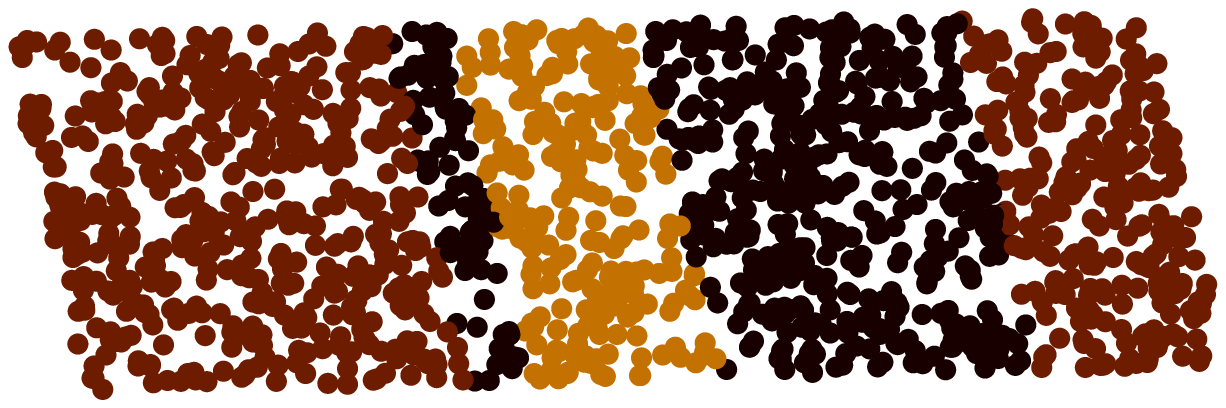}}
\caption{Left: Three classes laying on a swiss roll. Right: After unfolding them with MVU the classes are not linearly separable. Isometric Separation Maps managed to map this manifold in a 12-dimensional space such that the classes were linearly separable by 3 hyperplanes 100\% of the time and the 5-neighborhood distances were preserved with 0.1\% relative root mean square error}
\label{fig_3_class_cont_swiss_roll}
\end{figure}

In order to verify ISM on dimensionality
adjustment we tested it on the swiss roll dataset (1500 points). Two
classes were defined on the swiss roll that  were not linearly
separable. ISM was performed on the dataset. Embedding in 2
dimensions was not possible as the isometry cannot be preserved (the
algorithm terminated with 2\% error on  local distances). Embedding
was though possible in 3 dimensions where the algorithm terminated
with 0.01\% error in the local distance constraints. In both cases the
classification error was zero. As we see in
fig.~\ref{fig_two_class_swiss_roll} MVU unfolds the dataset in a
strip where the classes are not linearly separable. 
The ISM on the other hand transforms the manifold in a set that preserves the local
distances (k neighborhood=5) and divides the two classes in a
linearly separable way. In order to demonstrate further the power of
ISM we test it in two even more complex cases. In figure
\ref{fig_3_class_cont_swiss_roll} we generated 3 classes on a swiss
roll. Clearly MVU/MFNU unfolds the manifold in a non-separable way.
ISM was able to map the swiss roll in a 12-dimensional space where
the 3 classes are completely linearly separable. 

\begin{figure}[!h]
\centerline{\includegraphics[height=3.3cm]{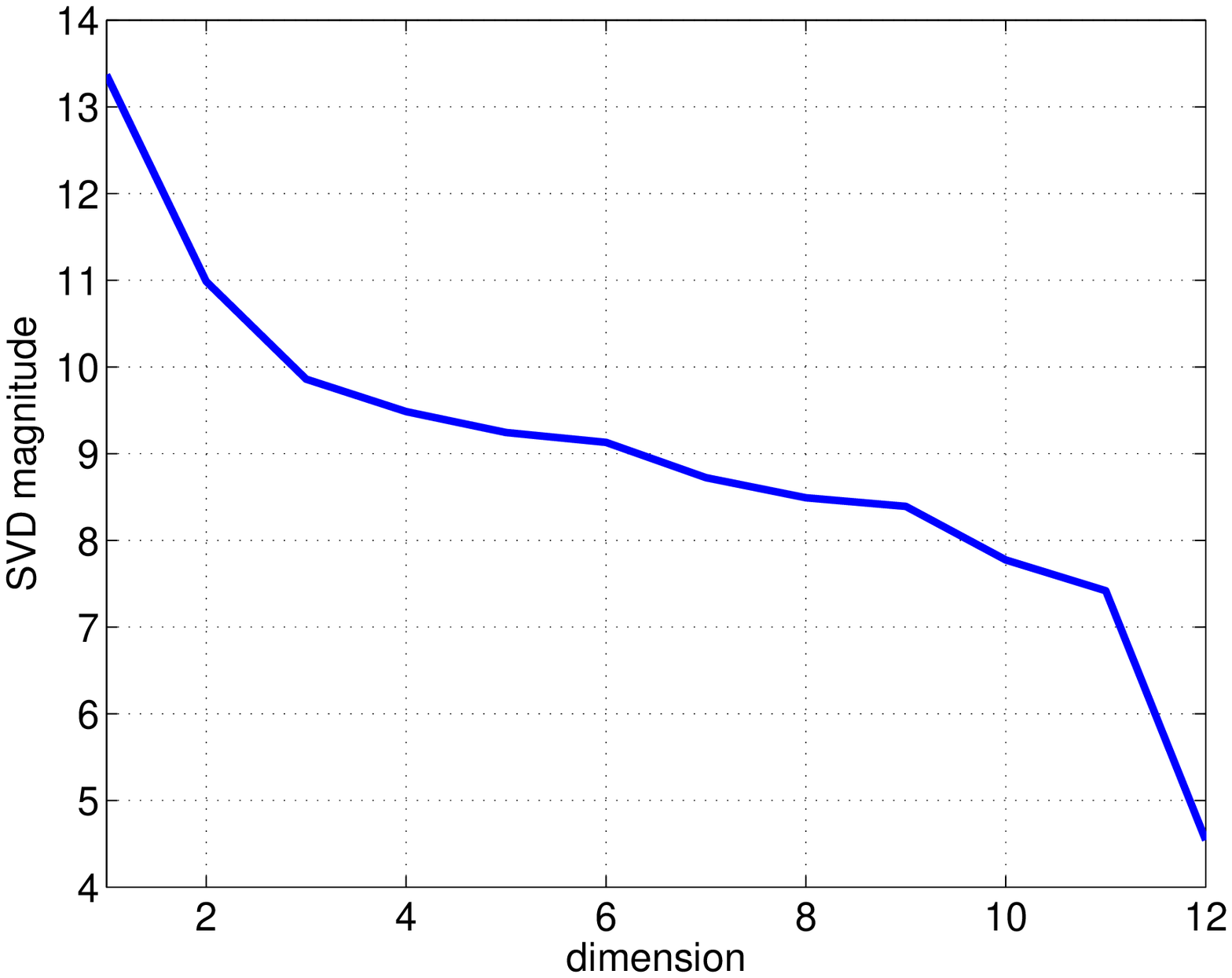}
            \includegraphics[height=3.3cm]{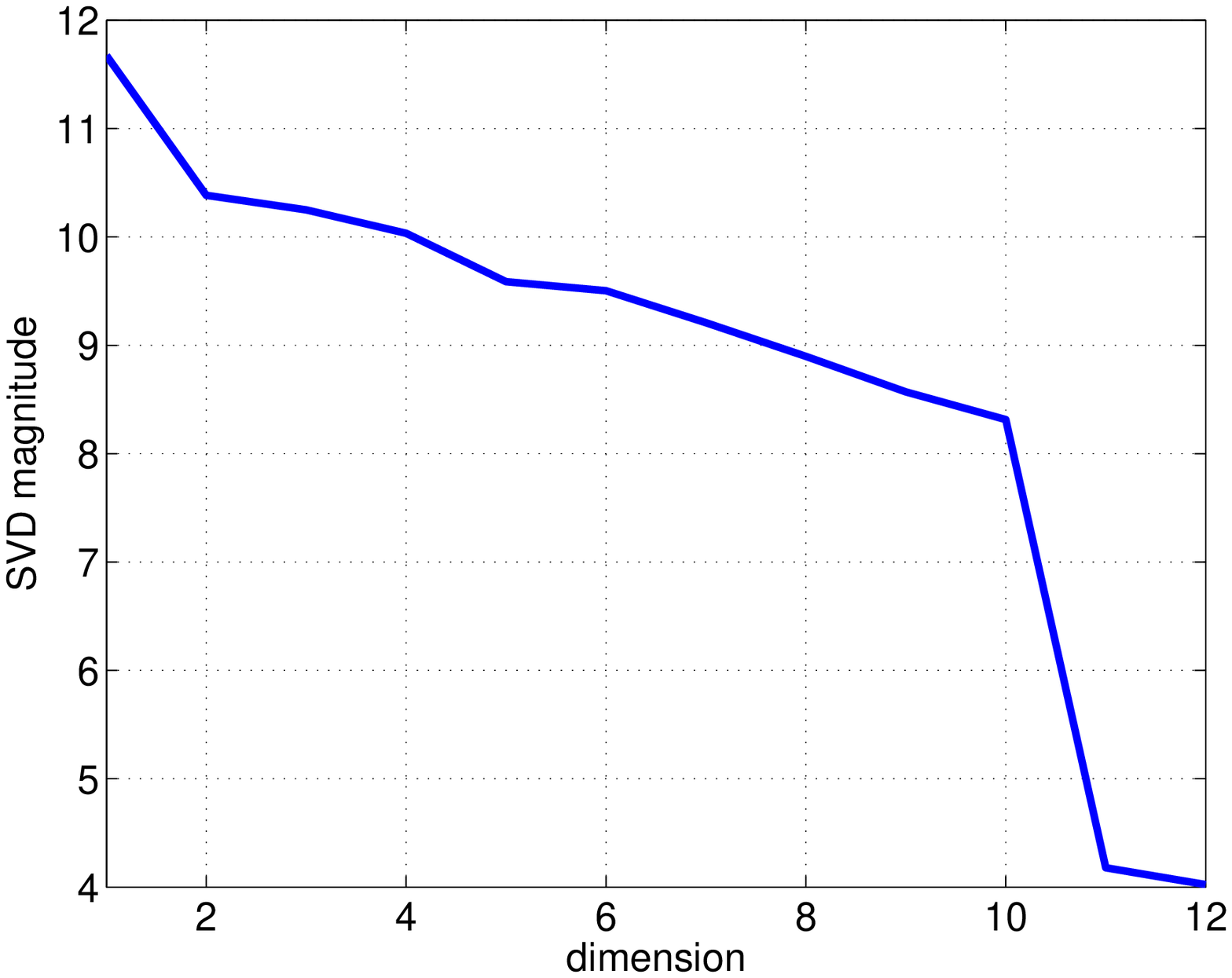}}
\caption{Left: We illustrate the PCA (SVD) spectrum of the unfolded swiss roll of figure \ref{fig_3_class_rand_swiss_roll}. As we can see it is pretty rich. Despite the bad structure of the classes, the ISM algorithm was able to map it on a 12 dimensional space. Right: We illustrate the PCA (SVD) spectrum of the unfolded swiss roll of figure \ref{fig_3_class_cont_swiss_roll}. As we can see it is pretty rich too.}
\label{fig_3_class_rand_cont_swiss_roll_svd}
\end{figure}

\begin{figure}[!h]
\centerline{\includegraphics[height=2.0cm]{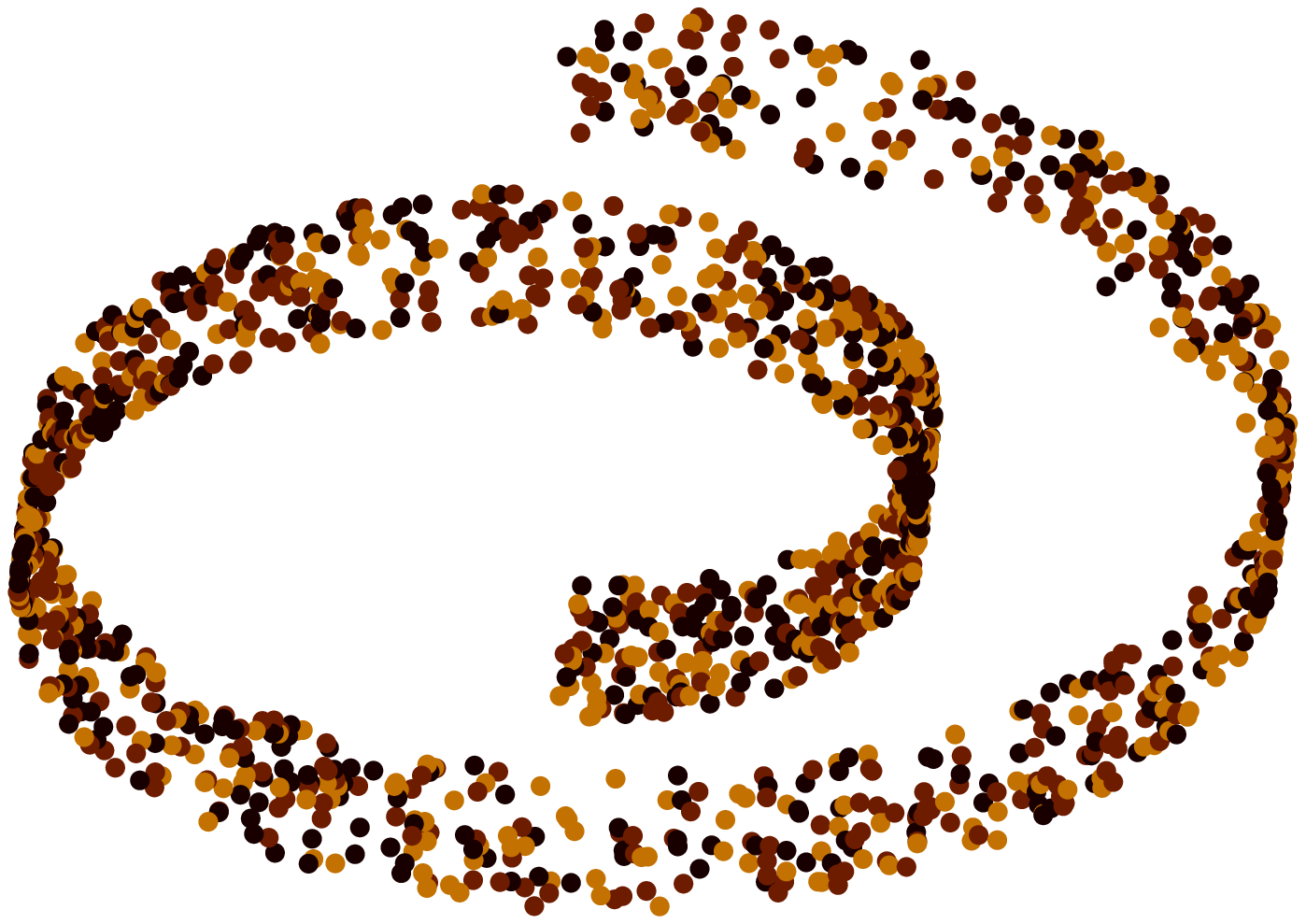}
            \includegraphics[height=2.0cm]{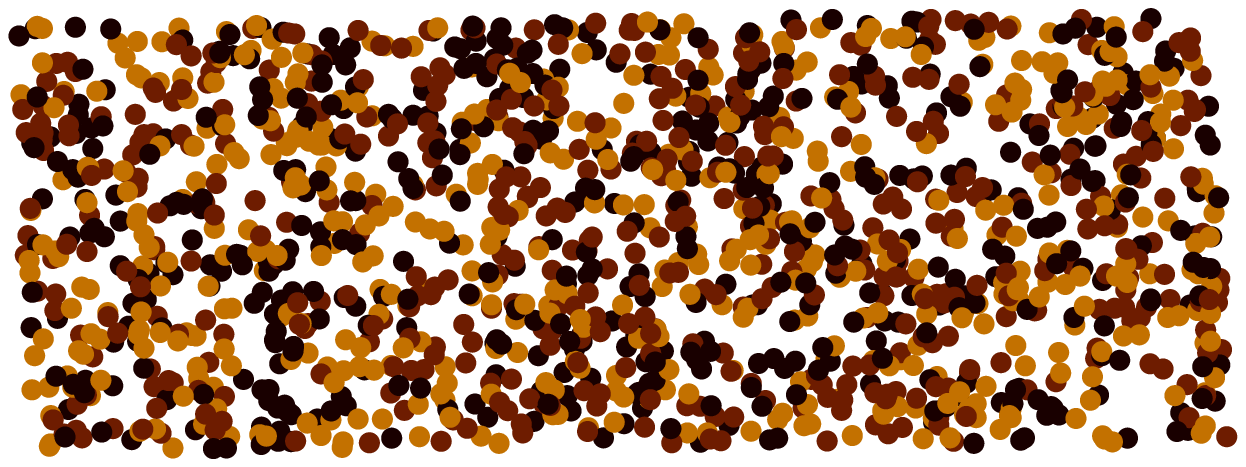}}
\caption{Top: Three classes laying randomly on a swiss roll. Bottom: After unfolding them with MVU the classes are not linearly separable. Isometric Separation Maps managed to map this manifold in a 12-dimensional space such that the classes were linearly separable by 3 hyperplanes. The optimization algorithm terminated with feasibility error 0.4\% for 5-neighborhood distance preservation, while 99.83\% of the points were correctly classified. The goal of this experiment was to verify experimentally that ISM can lift any strange dataset to a high dimensional space, such that classes are linearly separable}
\label{fig_3_class_rand_swiss_roll}
\end{figure}

In figure \ref{fig_3_class_rand_cont_swiss_roll_svd} the Principal Component
Analysis (PCA) spectrum of the 12 dimensional Swiss roll is shown.
The spectrum is quite rich. ISM can handle even more complicated
cases. In figure \ref{fig_3_class_rand_cont_swiss_roll_svd} we show 3 classes
lying randomly on a Swiss roll. ISM was able to map the manifold in
a 12-dimensional space keeping the 3 classes linearly separable. In
figure \ref{fig_3_class_rand_cont_swiss_roll_svd} the PCA spectrum is
depicted. The algorithm terminated with a very low feasibility error
0.4\% for distance preservation and 0.16\% for linear separability.
Further improvement of the feasibility error was possible, but
L-BFGS becomes slow as it goes close to the optimum. In general the
algorithm converges very quickly to 1\% feasibility error. Further
improvement is possible but takes time.

\section{Transductive SVMs}
\label{sec_TSVM} 
\begin{figure}[!h]
\centerline{\includegraphics[height=3.3cm]{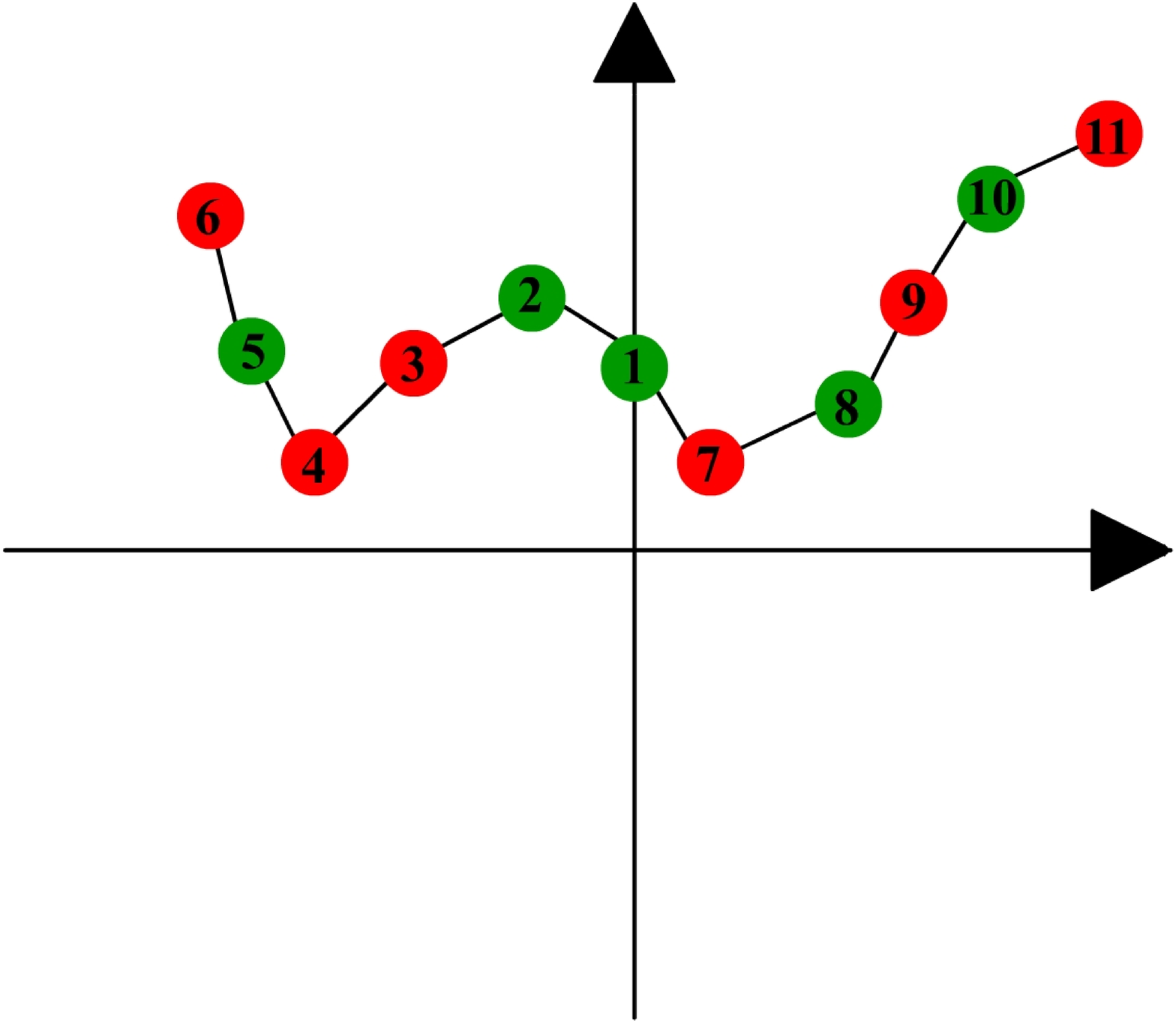}}
\centerline{\includegraphics[height=3.3cm]{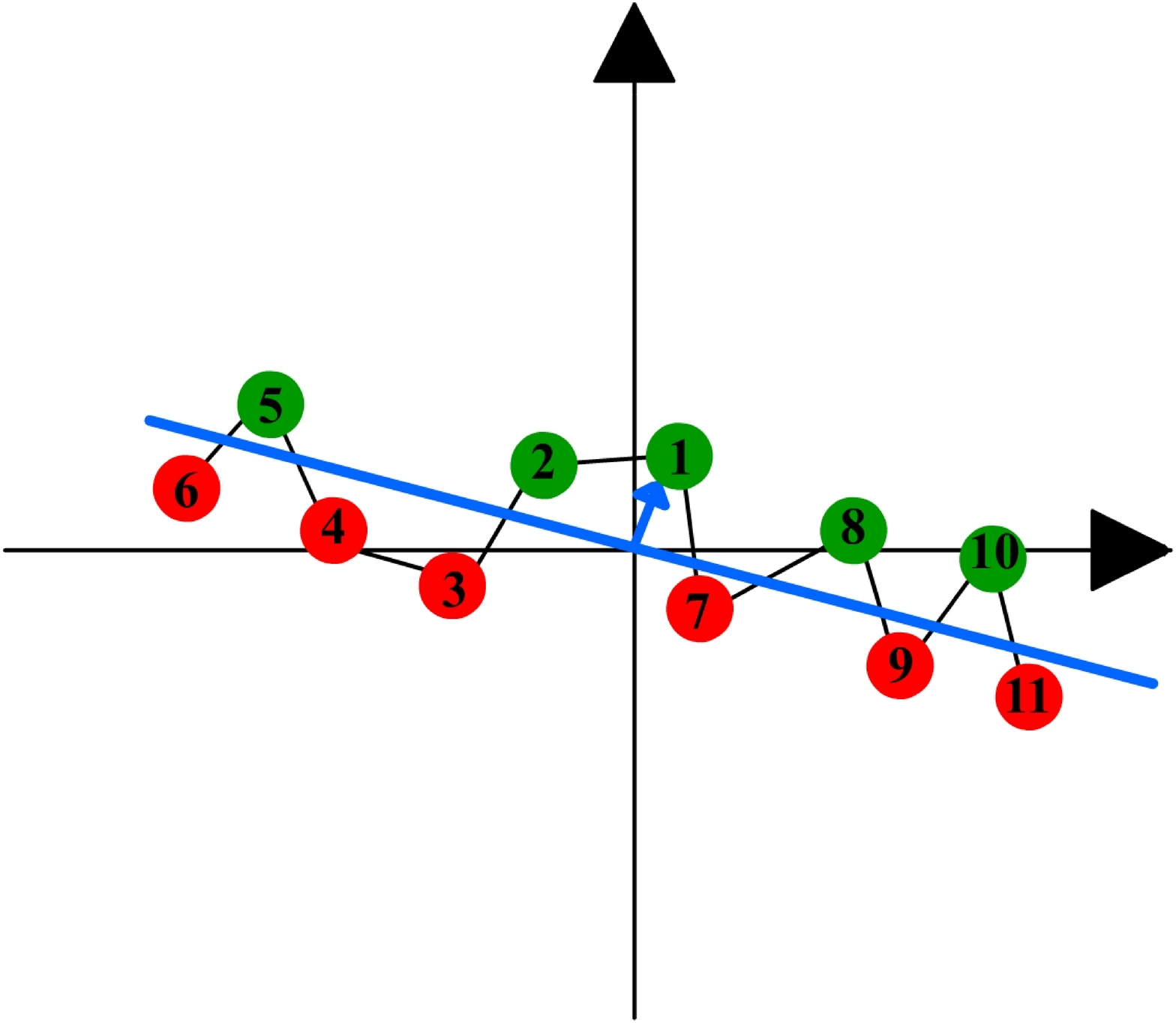}
           \includegraphics[height=3.3cm]{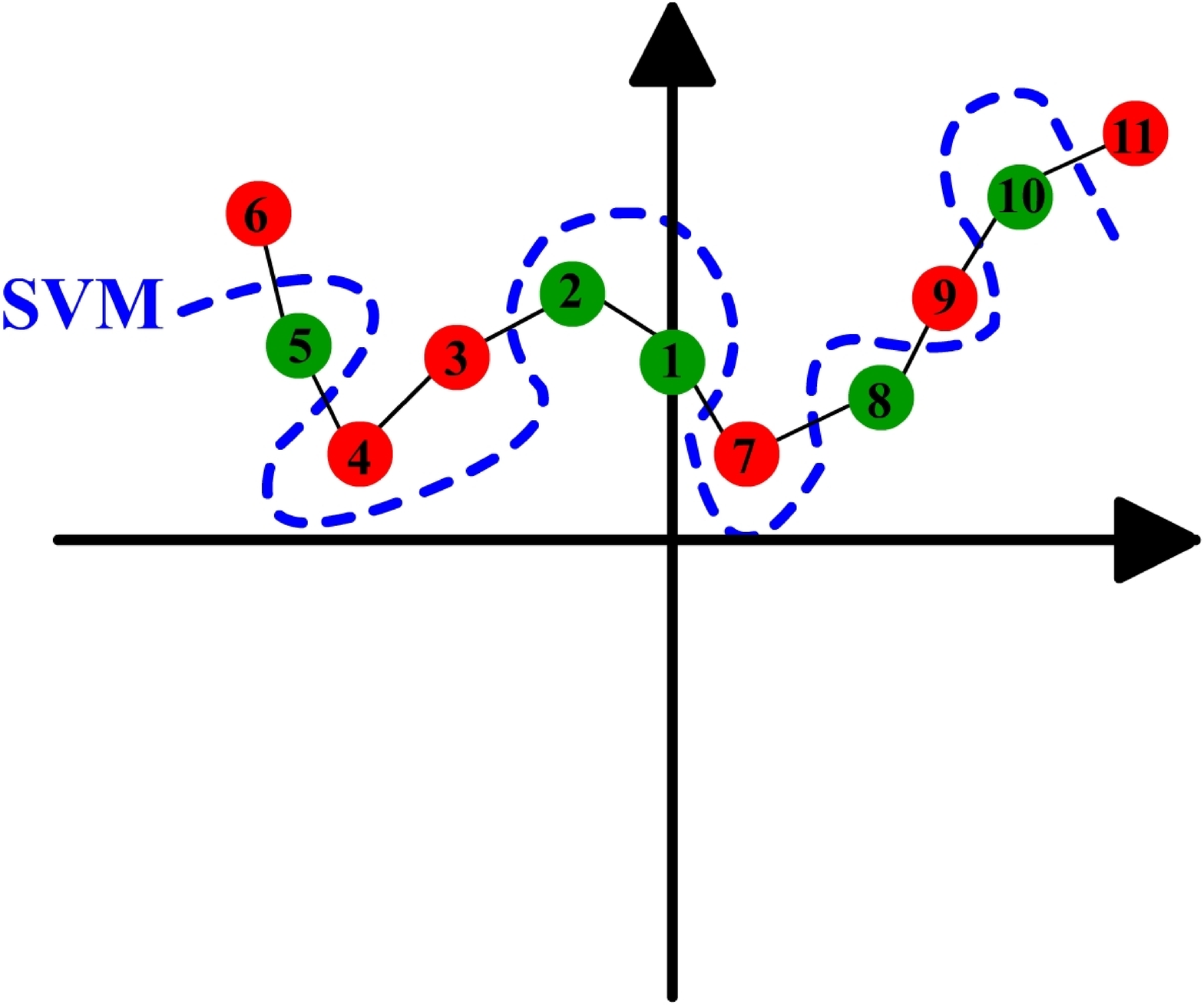}}
\caption{Top: A simple 2 class one dimensional manifold. Bottom:Left: ISM will pick a straight line and fold the points around it, so that it classes remain separable. Bottom:Right: The traditional SVM will keep the points fixed and find the curve that best separates the classes.}
\label{fig_ism_vs_svm}
\end{figure}

The method described above can also be used as a
transductive SVM in a semi-supervised setting. Transductive SVMs are
in general difficult problems. If the kernel is preselected then  a
mixed integer problem has to be solved. If the kernel is learnt from
the data then as we mentioned earlier no guarantee can be given that
the training data are linearly separable.  In ISM the kernel is
trained over all data, using all neighborhood information. After
solving the optimization problem, the classification information for
the test data will be on the sign of $K_{A,i} \; \forall i \in T$,
where $T$ is the test set. At this point we would like to highlight the difference between SVMs and ISMs. In figure~\ref{fig_ism_vs_svm} we see how SVMs and ISMs would classify points. SVMs keep the points fixed and try to find the optimal curve that separates the points. ISMs picks the hyperplane and moves the points around it (always keeping them connected),  so that they are correctly classified.

In order to evaluate ISM as an SVM classifier we chose a publicly
available  dataset and compared them versus traditional SVMs in two
different modes. We used the publicly available \textit{SVM-light}
software for traditional SVM classification. In the first experiment
we picked 1000 points from the magic gamma telescope dataset,
publicly available at the UCI repository. We chose 50 points as
training points and used the other 950 as test points. For
traditional SVM classification we tested the linear, Gaussian and
polynomial kernel, with different parameters for the the bandwidth
and the polynomial order. We also tuned the regularization factor so that the test error was minimized. In other words we pushed traditional SVMs to their best performance. The critical parameter for ISM SVM is the k-neighborhood. Usually small values of k allow embedding in lower
dimensional spaces, while large k lead in higher dimensional ones.
In tables \ref{ISM SVM1},\ref{T SVM1} the results are summarized. We
tested several k-neighborhoods for ISM and different Kernels for
traditional SVMs. From the results we observe that ISM behaves
slightly better that SVM (73.68\% versus 70.32\%). This is mainly because the training set is
small and SVM cannot capture very well the geometry.

In the second experiment we use the whole dataset. The training set
contains 12080 datapoints and the test set 6340. Although the
dataset is 10 dimensional, it is possible to reduce its dimension
with MVU/MFNU down to 5. In order to make it linearly separable
though with ISM it was necessary to use more than 10. In tables
\ref{ISM SVM2}, \ref{T SVM2} the results are summarized. As we can
see SVM performs slightly better than ISM (83.28\% versus 81.00\%). Another remark in
both cases is that ISM always behaves better than the linear kernel.
Gaussian SVM performance has the best performance. This is expected
since Gaussian matrices are usually full rank. ISM uses kernel
matrices of much smaller rank and they achieve equivalent
performance.

The results don't necessarily demonstrate big difference between
SVMs and ISM. We also experimented with some toy datasets, such as
the half moon dataset presented in \cite{belkin2005mr}and a Swiss roll, where one
data point is given per class. ISM obviously behaves better than SVM
but this is a trivial and not a fair comparison. In general the differences between ISM and traditional SVMs are in the same levels with the results reported in other transductive SVM papers \cite{belkin2005mr, lanckriet2004lkm}.In practice ISMs are slower than SVMs
since they are Semidefinite Problems contrary to SVMs that solve
Quadratic problems. It is interesting though that they provide an
tool for associating  the dimensionality of the dataset with the
classification score and linear separability. The more we increase
the dimensionality of the dataset with ISM the better the
classification score. In fact k acts as a regularizer. Large values
of k correspond to better generalization of the SVM as the test
error drops.

\begin{table}
\caption{ISM SVM Classification Score versus k-neighborhood for the
First Experiment} \label{ISM SVM1}
\begin{center}
{\small
\begin{tabular}{lll}
\multicolumn{1}{c}{\bf k-NEIGHBORS} & \multicolumn{1}{c}{\bf DIMENSION} & \multicolumn{1}{c}{\bf SCORE} \\
\hline \\
5  & 50 & 70.10\% \\
8  & 10 & 68.73\% \\
10 & 12 & 70.63\% \\
15 & 8  & 70.42\% \\
15 & 12 & 69.05\% \\
20 & 8  & 71.68\% \\
20 & 12  & 71.68\% \\
20 & 40  & 73.37 \\
25 & 8   & 72.63\% \\
25 & 12  & 71.89\%  \\
\textbf{30} & \textbf{40}  & \textbf{73.68\%} \\
\end{tabular}
}
\end{center}
\end{table}

\begin{table}
\caption{Traditional SVM Classification Score versus k-neighborhood}
\label{T SVM1}
\begin{center}
{\small
\begin{tabular}{lll}
\multicolumn{1}{c}{\bf KERNEL} & \multicolumn{1}{c}{\bf PARAMETER} & \multicolumn{1}{c}{\bf SCORE} \\
\hline \\
Gaussian   & 0.1 & 69.89\% \\
Gaussian   & 0.5 & 70.00\% \\
Gaussian   & 1.0 & 70.11\% \\
Gaussian   & 1.5 & 70.00\% \\
Gaussian   & 2.0 & 70.11\% \\
Gaussian   & 4.0 & 70.21\% \\
\textbf{Gaussian}   & \textbf{5.0} & \textbf{70.32\%} \\
Gaussian   & 6.0 & 70.21\% \\
Gaussian   & 8.0 & 70.11\% \\
linear     &  -  &  69.89\% \\
polynomial & 1   &  69.89\% \\
polynomial & 2   &  69.68\% \\
polynomial & 3   &  69.58\% \\
polynomial & 4   &  69.84\% \\
polynomial & 5   &  68.84\% \\
polynomial & 6   &  68.84\% \\
polynomial & 8   &  68.95\% \\
\end{tabular}
}
\end{center}
\end{table}

\begin{table}
\caption{ISM SVM Classification Score versus k-neighborhood For the
Whole Dataset} \label{ISM SVM2}
\begin{center}
{\small
\begin{tabular}{lll}
\multicolumn{1}{c}{\bf k-NEIGHBORS} & \multicolumn{1}{c}{\bf DIMENSION} & \multicolumn{1}{c}{\bf SCORE} \\
\hline \\
12 & 30 & 80.22\% \\
12 & 35 & 79.97\% \\
12 & 40 & 80.47\% \\
12 & 45 & 79.76\% \\
12 & 50 & 79.81\% \\
\textbf{12} & \textbf{55} & \textbf{81.00}\% \\
15 & 40 & 80.39\% \\
15 & 45 & 79.40\% \\
15 & 50 & 79.07\% \\
15 & 55 & 79.82\% \\
20 & 40  & 78.96\% \\
20 & 45  & 79.68\% \\
20 & 50  & 80.13\% \\
20 & 55  & 78.42\% \\
\end{tabular}
}
\end{center}
\end{table}

\begin{table}
\caption{ISM SVM Classification Score versus k-neighborhood For the
Whole Dataset} \label{T SVM2}
\begin{center}
{\small
\begin{tabular}{lll}
\multicolumn{1}{c}{\bf k-NEIGHBORS} & \multicolumn{1}{c}{\bf DIMENSION} & \multicolumn{1}{c}{\bf SCORE} \\
\hline \\
\textbf{Gaussian}   & \textbf{8} & \textbf{83.28\%} \\
Gaussian   & 6 & 82.77\% \\
linear     & - & 78.64\% \\
polynomial & 2 & 81.62\% \\
polynomial & 3 & 82.07\% \\
polynomial & 5 & 81.26\% \\
\end{tabular}
}
\end{center}
\end{table}

\section{Summary}
In this paper we presented a new Manifold Learning method the
Isometric Separation Maps. This method is ideal for reducing the
dimension of Manifold with class information associated with them.
We also showed how ISM can be used as semi-supervised (transductive)
classifiers. Although they don't have superior performance compared
to traditional max margin SVMs, they are a useful tool for
determining the dimensionality of the kernel space that is necessary
for achieving linear separability. We believe that some improvement of the objective function is necessary so that generalization is improved. Probably a term minimizing the norm of the vector normal to the hyperplane (as in SVMs) can be used.


\bibliographystyle{IEEEbib}
\bibliography{mlsp2009}

\end{document}